\pdfoutput=1
\documentclass[11pt]{article}

\usepackage[final]{acl}

\usepackage{times}
\usepackage{latexsym}

\usepackage[T1]{fontenc}
\usepackage[utf8]{inputenc}

\usepackage{microtype}

\usepackage{inconsolata}

\usepackage{graphicx}
\usepackage{booktabs}
\usepackage{soul,color,xcolor}
\usepackage{amsmath,amsfonts}
\usepackage{bbding}

\definecolor{delectricblue}{RGB}{255,182,193}
\colorlet{lightdelectricblue}{delectricblue!30}
\newcommand{\ch}{\color[HTML]{5BCB13}}
\newcommand{\x}{\color[HTML]{CB0505}}

\title{\includegraphics[width=1.0cm]{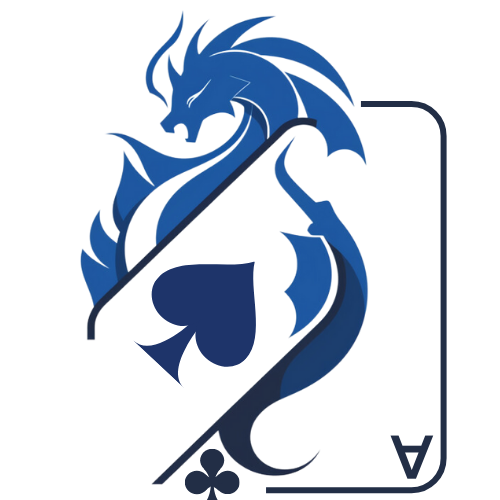}Instruction-Driven Game Engine: A Poker Case Study}

\author{Hongqiu Wu\textsuperscript{\rm 1,2,3}$^\dag$ \and Xingyuan Liu\textsuperscript{\rm 1,2,3}$^\dag$ \and Yan Wang\textsuperscript{\rm 4}$^*$ \and Hai Zhao\textsuperscript{\rm 1,2,3}\thanks{Corresponding author. $^\dag$ Equal contribution. This research was supported by the Joint Research Project of Yangtze River Delta Science and Technology Innovation Community (No. 2022CSJGG1400),
the Joint Funds of the National Natural Science Foundation of China (Grant No. U21B2020).} \\
$^1$Department of Computer Science and Engineering, Shanghai Jiao Tong University\\
$^2$Key Laboratory of Shanghai Education Commission for Intelligent Interaction \\
and Cognitive Engineering, Shanghai Jiao Tong University \\
$^3$Shanghai Key Laboratory of Trusted Data Circulation and Governance in Web3 \\
$^4$Tencent \\
\texttt{\{wuhongqiu,chloelxy,zhaohai\}@sjtu.edu.cn,yanwang.branden@gmail.com}}

\begin{document}

\maketitle
\begin{abstract}
The \textbf{\textit{Instruction-Driven Game Engine (IDGE)}} project aims to democratize game development by enabling a large language model (LLM) to follow free-form game descriptions and generate game-play processes. The IDGE allows users to create games simply by natural language instructions, which significantly lowers the barrier for game development. We approach the learning process for IDGEs as a \textit{Next State Prediction} task, wherein the model autoregressively predicts the game states given player actions. The computation of game states must be precise; otherwise, slight errors could corrupt the game-play experience. This is challenging because of the gap between stability and diversity. To address this, we train the IDGE in a curriculum manner that progressively increases its exposure to complex scenarios.
Our initial progress lies in developing an IDGE for Poker, which not only supports a wide range of poker variants but also allows for highly individualized new poker games through natural language inputs. This work lays the groundwork for future advancements in transforming how games are created and played.
\end{abstract}

\section{Introduction}

\begin{figure*}
\centering
\includegraphics[width=0.85\textwidth]{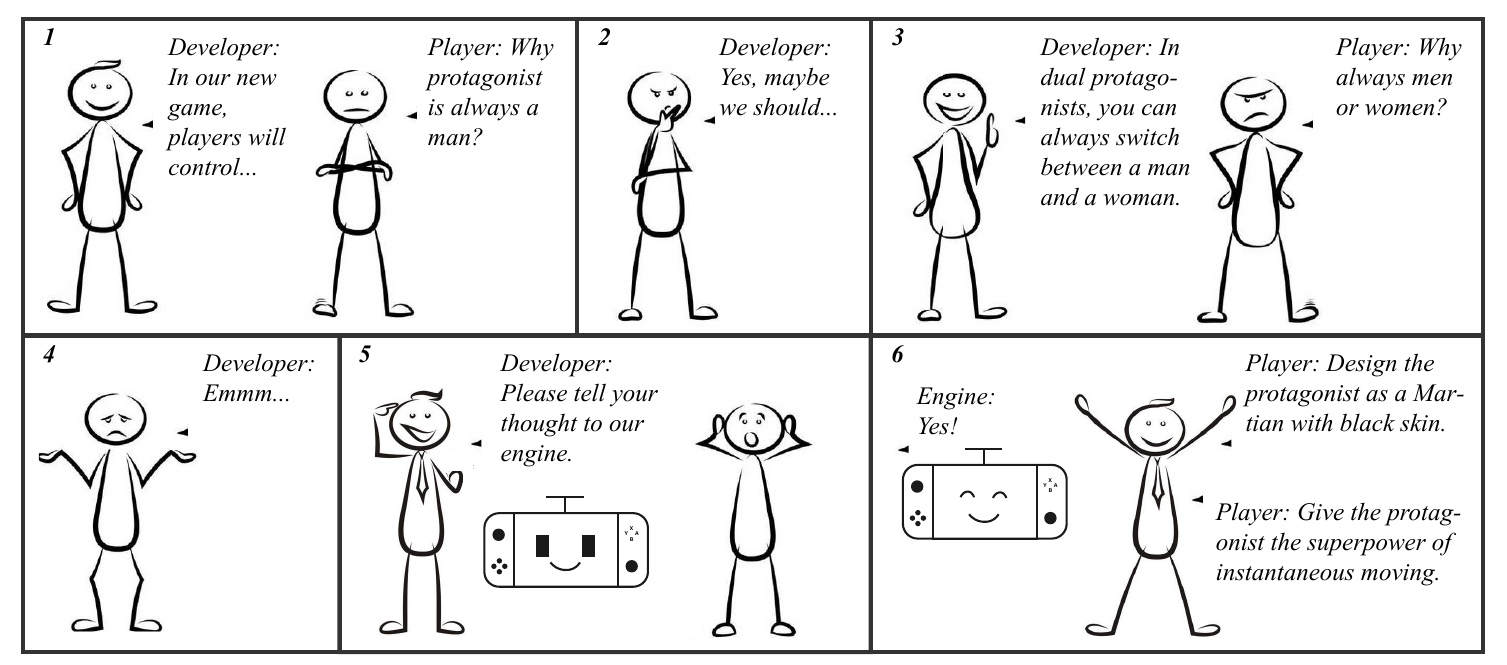}
\caption{1: Players were tired of the game's protagonist models. 2, 3: Developers thus created a new mode with dual protagonists. Players still didn't buy it, while they didn't know how to develop games. 4: There were irreconcilable divides between players and developers. 5, 6: Till the advent of the IDGE, it can read the players' mind and let them experience the games immediately.}
\label{fig:comic}
\end{figure*}

Game developers dedicate creativity to offer immersive experiences to game players. Players immerse themselves in games and offer valuable feedback to developers. This makes a symbiotic relationship between creators and customers.
However, as depicted in the comic from Figure~\ref{fig:comic}, there are disconnections between them, due to diverse preferences of players across age, gender, and cultural backgrounds. Despite the fact that many today's games allow for customization of basic characters and appearances, it is an impossible task for developers to craft every aspect of the game to suit the need of every player. Our study seeks to reconcile such a divide.

Game engines, as the heart of game development, are conventionally driven by programming languages. This technical barrier often deters enthusiasts from realizing their game development dreams. In response, we propose a novel concept: \emph{Instruction-Driven Game Engine (IDGE)}, a game engine enabling anyone to fashion a game through natural language instructions and generating the resultant game-play process.
Distinct from recent advancements in video-based games \citep{DBLP:conf/icml/BruceDEPS0LMSAA24,DBLP:journals/corr/abs-2404-10179}, our focus in this paper is on the text-based game states. We leverage Unity to render these states to visual display.

IDGE is a neural engine, meaning it is built upon neural networks, specifically large language models (LLMs) \citep{DBLP:conf/nips/BrownMRSKDNSSAA20,DBLP:journals/corr/abs-2303-08774,DBLP:journals/corr/abs-2307-09288,DBLP:journals/corr/abs-2309-10305}. It is designed to follow a \textbf{game script}, a detailed instruction that blueprints the game, e.g. settings, rules, elements, and drive the progression of game-play as interacting with players. IDGEs frame the operation of engines as a \textit{Next State Prediction} task, which autoregressively predicts the next game state based on the user-specified game script, previous game state, and current player action.

Training an IDGE faces the dual challenges of \textbf{stability} and \textbf{diversity}. The former seeks to provide a stable and precise game-play throughout lengthy contexts, while the latter seeks to follow diverse preferences across the large player base. Unfortunately, we empirically see an ironic twist: the model trained directly from naive game logs is neither stable nor diverse. Therefore, we employ a standard-to-diverse curriculum learning methodology, which gradually introduces complexity into the training process, incrementally enhancing the model's diversity while preserving its stability.

While it is still on journey from building an IDGE capable of producing AAA games, this paper provides an initial progress on \textbf{Poker}, a worldwide card game, e.g. \emph{Texas hold'em}, \emph{Badugi}.
We train the IDGE using data sourced from a poker simulator.
We show that the IDGE's understanding of nuanced semantics successfully fills voids left by the simulator program, e.g. generating suits and numbers that never occurred in the training process.
Furthermore, the IDGE shows immense promise in generalizing to entirely new games, e.g. handling novel card combinations and battle strategies.

We summarize our paper below:
$\bullet$ $\S$ \ref{s2} introduces the concept of the IDGE and its learning problem;
$\bullet$ $\S$ \ref{s3} discusses the IDGE-style data for poker games;
$\bullet$ $\S$ \ref{s4} proposes the enhanced training techniques.

\section{Instruction-Driven Game Engine}
\label{s2}

In this section, we introduce dialogue-style LLMs as the setup for \textit{IDGEs}. We then formulate the learning problem as \textit{Next State Prediction}.

\subsection{From Instruction-Driven Dialogue to Instruction-Driven Game Engine}
Most LLMs \citep{DBLP:conf/nips/BrownMRSKDNSSAA20,DBLP:journals/corr/abs-2303-08774,DBLP:journals/corr/abs-2307-09288,DBLP:journals/corr/abs-2309-10305} have been fine-tuned on dialogue-style corpora, where it is endowed with the ability to interact with users.
The resultant models can follow a system instruction provided by users and lead to a dialogue process in line with it.

Likewise, an IDGE works through interaction, too. Its system instruction specifically refers to a game script that accurately describes the desired game. In game-play, the IDGE interacts with players (users), concurrently processing player inputs, (e.g. moves, targets), to dynamically generate the game states as responses.

\begin{figure*}[t]
\centering
\includegraphics[width=0.86\textwidth]{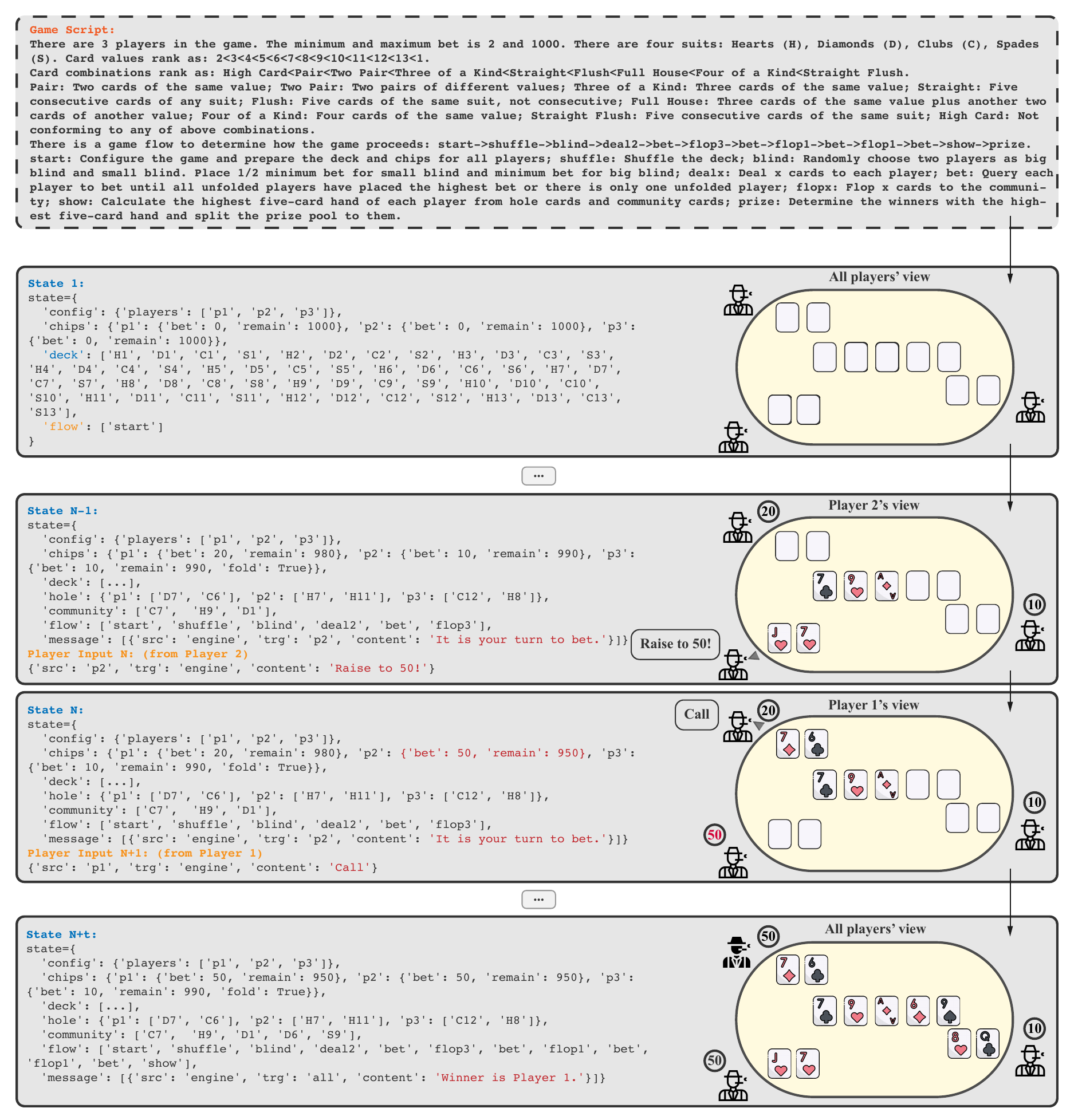}
\caption{Game-play samples for next state prediction. In the lower half, we illustrate the state prediction circle using NSP. The left side is the input text for the engine from a global view, including all parts that are visible to players as well as those that are not. The right side is the diagram of the game from different players' views.}
\label{fig:ex}
\end{figure*}

In Figure~\ref{fig:ex}, we demonstrate how a poker IDGE facilitates a variant of \textit{Texas Hold'em}: the player first inputs the game script in natural language.
Based on this game script, the IDGE simulates the game-play process with the player state by state. The player performs the action, e.g. check, call, raise, and the engine computes and returns the resultant game state. It is a dialogue-like process and will continue till the game concludes.

\subsection{Next State Prediction}

Causal language models learn the interplay of words through the autoregressive process of next token prediction \citep{DBLP:conf/nips/VaswaniSPUJGKP17,DBLP:conf/nips/BrownMRSKDNSSAA20}.
From a game-play perspective, the minimum component is no single token, but rather each \textbf{game state}.
A game state is a single frame that contains all real-time game information, e.g. characters, items, missions. Essentially, the task of any game engines is exactly to compute the next state according to the prior ones. Therefore, we may formulate the learning of IDGEs as a \emph{Next State Prediction (NSP)} problem.

Given a sequence of game states $\mathbf s=\{s_0,s_1,\cdots,s_T\}$, an IDGE with parameters $\theta$ seeks to maximize the likelihood:
\begin{equation}
    \sum_{t=1}^{T}\log p_\theta(s_t|s_0,s_1,\cdots,s_{t-1},x_t,z)
    \label{e1}
\end{equation}
where $x_t$ refers to the player input at the moment $t$ and $z$ refers to the game script which is global for the entire game.
The engine seeks to predict the next state $s_t$ given the prior states $s_0,s_1,\cdots,s_{t-1}$ following $z$.

A game state is typically far bigger than a token, incurring overflow of inputs and posing challenges for language models in capturing long-range dependencies \citep{DBLP:journals/corr/abs-2004-05150,DBLP:conf/iclr/XiaoTCHL24}.
A more manageable case occurs when it is assumed that each state $s_t$ solely depends on its previous $k$ states.
Specifically when $k=1$, Eq. \ref{e1} can be reduced to:
\begin{equation}
    \sum_{t=1}^{T}\log p_\theta(s_t|s_{t-1},x_t,z).
    \label{e2}
\end{equation}
While such an independence assumption would incur information loss, a solution is to keep a summary module within the game state.

NSP is a general way to model the process of game-play using a neural engine. However, the practical performance will be limited by models' computational capabilities. For example, it won't be easy for an LLM to handle sophisticated numerical calculation \citep{DBLP:conf/emnlp/Wu0ZZ23}, especially for a smaller one.
To overcome this weakness, we augment the state prediction process using code modality. The engine is allowed to predict the intermediate code to serve its duty rather than offering the eventual results directly. The prediction will be post-processed by a code interpreter to compute the next state eventually.
A toy example is the shuffling of poker cards. It is very hard for a neural model to generate uniformly distributed cards from its inner representation. To do this, it can define a ``shuffle'' function and then call it in the next state.

In addition to defining new functions or methods, we allow the engine to call predefined functions, called \textbf{core functions}, which are defined in an external core set.
These core functions are usually the essential routines that will be frequently used in the game, such as shuffling, ranking of cards in poker.
By utilizing core functions, the engine further overcomes the inefficiency of generating repeated content.

The integration of core functions extends IDGEs' functionality and flexibility, enabling them to handle a broader scope of games.
This design is akin to the hierarchical architecture in conventional game engines, where the high layers are allowed to call utilities from the core layer.

\subsection{Differential State Prediction}
The inference complexity of NSP scales quadratically with the sequence length. Therefore, decoding a lengthy game state may fall into trouble.
Empirically, the game state only undergoes a slight change between two successive moments $t$ and $t+1$, with the majority of the state remaining the same. This phenomenon can be potentially general across various games when the intervals between states are short.
We thus introduce \emph{Differential State Prediction (DSP)}, an efficient variant of NSP, where the engine is simplified to predict solely the difference of two states:
\begin{equation}
    \sum_{t=1}^{T}\log p_\theta(\Delta s_t|s_{t-1},x_t,z)
    \label{e3}
\end{equation}
where $\Delta s_t$ is the difference of $s_{t-1}$ and $s_t$.
DSP is more efficient compared to NSP in most situations, significantly accelerating the inference during game-play.
In our experiments, we find that DSP also produces slightly better performance.

To reconstruct $s_t$ from $\Delta s_t$ and $s_{t-1}$, there will be a merge function $s_t=M(\Delta s_t,s_{t-1})$.
In this work, each game state is implemented as a dict. Hence, $M$ refers to the coding of updating dict elements.
The following section will demonstrate concrete examples of NSP/DSP for a poker game.

\begin{figure}[t]
\centering
\includegraphics[width=0.49\textwidth]{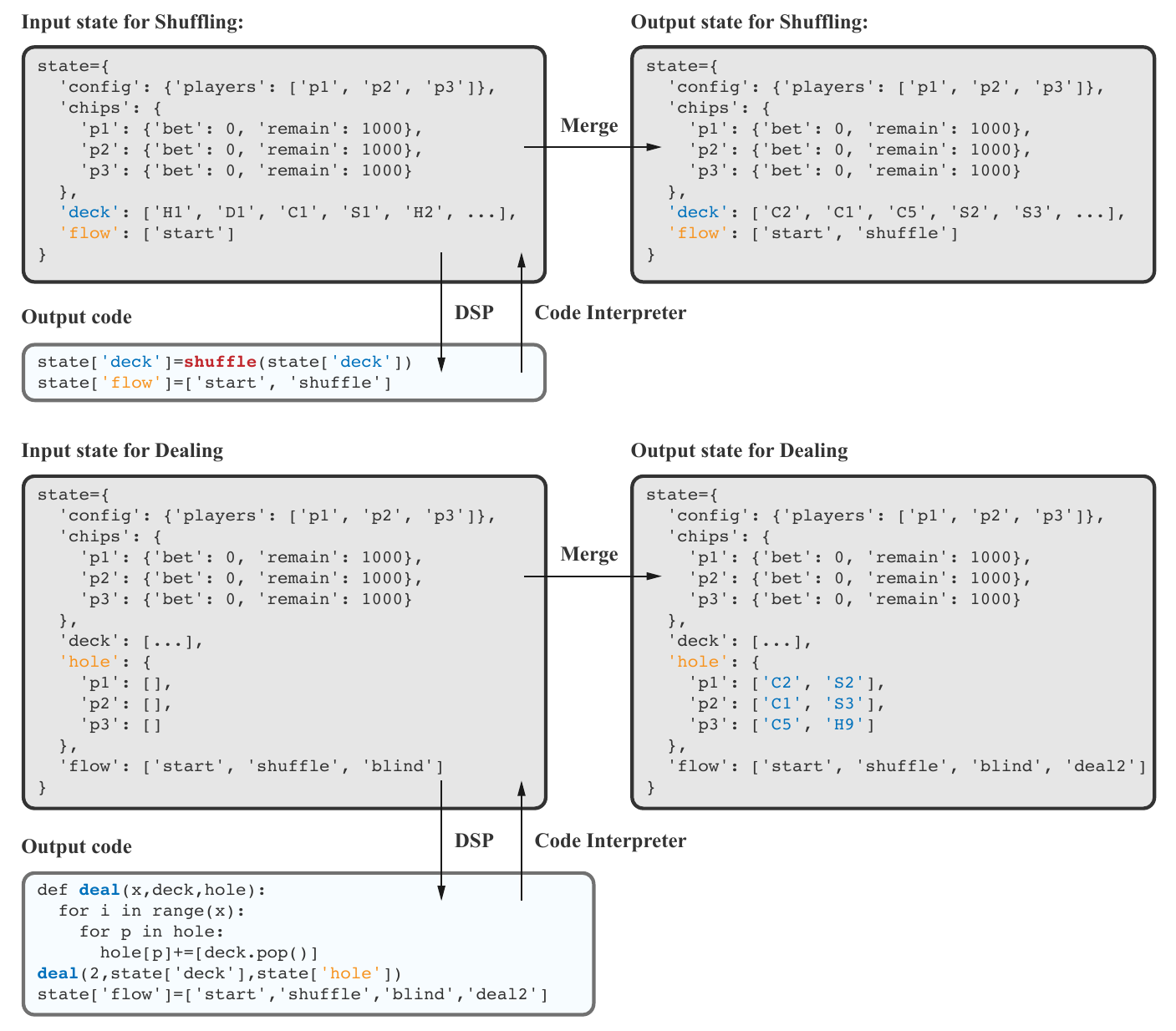}
\caption{DSP. In the shuffling case, the IDGE calls ``shuffle'', which is a predefined core function. In the dealing case, it defines a new ``deal'' function to deal a number of cards to each player one by one. We use a code interpreter to merge the input state and the output code to obtain the next state.}
\label{fig:dsp}
\end{figure}

\section{Data for IDGE}
\label{s3}

Our training data is sourced from two methods. First, we develop a poker simulator and obtain the training data from its game logs. The simulator supports ten representative poker games: \textit{Texas Hold'em}, \textit{Omaha}, \textit{Omaha HL}, \textit{Short-deck Hold'em}, \textit{2-to-7 triple Draw}, \textit{A-to-5 triple Draw}, \textit{2-to-7 single Draw}, \textit{Badugi}, \textit{Badeucey}, and \textit{Badacey}. Additionally, it allows for further configuration of several common poker elements, e.g. type of suits, numbers. By adjusting these elements, one can derive virtually infinite variations beyond aforementioned ten poker games.
Moreover, we realize that if the game logs are sampled completely in uniform, the occurrence of some rare states, such as some superior card combinations, would be extremely low. The resultant engine trained on such data may fall short in low-frequency situations, even though the dataset is large. Therefore, we balance the data by up/down-sampling the game logs to ensure that all possible situations occur similarly.
After obtaining game logs, we transform each log into a training sample as in Figure \ref{fig:ex} for NSP and DSP. Each sample is made up of three parts: the game script $z$, player input $x_t$, and game states $s_t$. If we were to draw an analogy with ChatGPT, they respectively play the roles of the system, user, and assistant.

% The histograms in Figure \ref{fig:cb} provide an instance where we balance the card combinations within training data.
% Typically in a poker game, the chance of a straight (e.g. 5, 6, 7, 8, 9) in hand is much lower than a pair (e.g. 6, 6).
% As depicted in Figure \ref{fig:cb} (left), recorded from 1,000 poker games, we find that the occurrences of ``pair'' and ``high card'' far outstrip the others.
% We thus manipulate the data sampling process so that each combination occurs similarly, as in Figure \ref{fig:cb} (right).
% We find that such a simple manipulation greatly enhances the training outcome with a smaller amount of training data.

The second part of the data is generated by GPT3.5. Based on the state prediction data from the simulator, we prompt GPT3.5 to augment the game scripts and generate the corresponding new game states. This process is manually done by skilled prompting, which incorporates scenarios beyond typical poker games, expanding the diversity of training data as a result.

\noindent$\spadesuit$ \textbf{Game Script}\;
To describe the poker game in natural language, we design a prototype game script. The top part of Figure~\ref{fig:ex} illustrates the game script for a \emph{Texas Hold'em} variant.
We can see that it defines a series of game elements: the number of players in the game, minimum and maximum bet limits, suits and values of single cards, battle strategies, and the game flow. These elements correspond to the configuration of the poker simulator.
Particularly, the game flow refers to the procedures this game will go through in order, e.g. bet, deal.

\noindent$\spadesuit$ \textbf{Game State and Player Input}\;
For the game state and player input, we adopt a dict format as shown on the left side of Figure \ref{fig:ex}. For instance, ``deck'' is followed by the remaining cards in the deck, ``hole'' and ``community'' is followed by the hole cards of players and the public cards, while ``message'' is followed by the message sent from the source $a$ to target $b$.
On the right side of Figure \ref{fig:ex}, we show the diagram of the poker game associated to the left-side game state. In player input $N$, player 2 chooses to raise the bet. Given state $N-1$, the engine outputs state $N$, where the chips of player 2 are updated and player 1 is informed to bet since player 3 has folded.

To ensure the independence assumption that each state $s_t$ solely depends on state $s_{t-1}$, we incorporate the game flow as a summary module in the game state. It specifically caches all past game procedures in order. The engine can thus be navigated to step into the next procedure correctly, regardless of the amount of game-play history.

\begin{table}[]
\centering
\small
\begin{tabular}{rrrrr}
num. of  & len. of & len. of & len. of output             & avg.        \\
samples  & script  & input   & NSP $\rightarrow$ DSP      & states      \\ \toprule
10k      & 439.8   & 401.1   & 404.9 $\rightarrow$ 135.9  & 35.3        \\
\end{tabular}
\caption{Statistics of training data.}
\label{t:dataset}
\end{table}

\noindent\textbf{Data Statistics}\;
Table \ref{t:dataset} shows the statistics of the training data that we construct, which comprises 10k state prediction samples. 
Specifically, the average number of states of one game is 35.3, i.e. the number of states for the engine to predict. The output tokens of DSP is much less than that of NSP.

\section{Curriculum Learning}
\label{s4}

Straightforwardly, we could utilize the data generated in $\S$ \ref{s3} to fine-tune a base model by maximizing Eq. \ref{e2}/\ref{e3} and obtain the IDGE. However, the resultant IDGE may struggle with stability and diversity: neither can it accurately predict the next game state nor comprehend the user-specified game script in natural language. Therefore, we devise a progressive curriculum learning process \citep{DBLP:conf/icml/BengioLCW09}, to incrementally enhance the IDGE's diversity while preserving stability.

\noindent\textbf{Warmup: Training on Core Set}\;
In $\S$ \ref{s2}, we utilize a set of core functions to facilitate the process of state prediction. Though the model can be exposed to all core functions via fine-tuning, we observe that it struggles to call the core functions properly. This phenomenon is much more severe in unseen contexts. We attribute this to the cold start problem that the model merely memorizes the names of the core functions during training, without knowing their underlying implementation.
To this end, we introduce a pre-learning phase to warmup the model.
We develop an instruction tuning dataset of 1k samples derived from the core set, where each core function is translated to a natural language instruction and the model is trained to implement the function in a way of instruction following.
This phase offers a profound comprehension of the model's usage of core functions.

\noindent\textbf{Standard: Training on Standard Game Scripts}\;
The next step is to train the model on the standard data introduced in $\S$ \ref{s3} by optimizing NSP/DSP.
In this phase, the model is forged into an engine, predicting game-play state by state following the game scripts, and is combined with pre-learned core functions organically.

\noindent\textbf{Diverse: Training on Rephrased Game Scripts}\;
While the standard data already includes the prototype game scripts, mastering the prototype descriptions can be too restrictive for users. Rather, it is more natural for them to describe their desired games in free-form natural language.
Rather than exhaustively crafting new natural language data, we introduce \emph{Segment Rephrasing (SR)}, a technique that rephrases a portion of the game script to encourage the model to follow diverse natural language.
Specifically, given a game script, we segment it into chunks and randomly rephrase several of them.
To largely keep the semantics intact, there is only a very low probability that the entire script will be rephrased.
The rephrasing process is done by GPT3.5.
These rephrased game scripts enable the model fully ``to customers''. In addition, these scripts will be more challenging to understand, which potentially generalizes the model to unseen scenarios.
Readers may refer to Table \ref{tab:oo-d-script} in Appendix \ref{apx:2} for real human-written examples.

We summarize the training pipeline for the IDGE:
1) train on the core set $\mathcal D_{cs}$ (1k); 2) train by optimizing NSP/DSP on the standard dataset $\mathcal D$ (10k); 3) rephrase the standard data $\mathcal D_{sr}$ and train on the sum of $\mathcal D$ and $\mathcal D_{sr}$ (20k).

The \textbf{warmup}, \textbf{standard}, and \textbf{diverse} process correspond to the easy, medium, and hard curriculum.
It serves for a smooth transfer of the IDGE from standardization to diversity.

\section{Experimental Results}
\label{s6}

In this section, we evaluate the IDGE in two scenarios. The former is automatically generated by our simulator, which can be considered as a test set that has the same distribution as the training set. The latter resembles the real-world situations, where proficient poker players are directly enlisted as annotators to create new game scripts. Subsequently, the test data is obtained by playing the games online by themselves with the IDGE.

\subsection{Training and Evaluation Setup}
We develop the IDGE based on CodeGemma-7b \citep{team2024gemma}\footnote{\url{https://huggingface.co/google/codegemma-7b-it}}. CodeGemma is a code LLM that is additionally pre-trained on large code corpora. We find that CodeGemma works better than similar-sized natural language models like LLaMA3 \citep{DBLP:journals/corr/abs-2407-21783}.
We train each model using LoRA \citep{DBLP:conf/iclr/HuSWALWWC22} with $r=8$, $\alpha=32$, and the optimal learning rate in 1.5e-4 and 3e-4.
The warmup of the learning rate is set to 30 steps and the total batch size is set to 8 on 8 chips.
For each curriculum, we train 3 epochs.
To ensure the stability of outputs, we leverage greedy decoding.

\begin{table*}[t]
\centering
\small
\setlength\tabcolsep{5.3pt}
\begin{tabular}{|l|c|c|c|c|c|c|c|c|c|c|}
\hline
                      & \textit{Texas} & \textit{Omaha} & \textit{Om. HL} & \textit{Short.} & \textit{27 triple} & \textit{A5 triple} & \textit{27 single} & \textit{Badugi} & \textit{Badeucey} & \textit{Badacey} \\ \hline
NSP                   & \ch\checkmark  & \ch\checkmark  & 18/20             & \ch\checkmark       & 18/20                  & \ch\checkmark          & \ch\checkmark          & \ch\checkmark   & 17/20             & 18/20            \\ \hline
DSP                   & \ch\checkmark  & \ch\checkmark  & \ch\checkmark     & \ch\checkmark       & \ch\checkmark          & \ch\checkmark          & \ch\checkmark          & \ch\checkmark   & 18/20             & 18/20            \\ \hline
DSP (CS)              & \ch\checkmark  & \ch\checkmark  & \ch\checkmark     & \ch\checkmark       & \ch\checkmark          & 19/20                  & \ch\checkmark          & \ch\checkmark   & \ch\checkmark     & \ch\checkmark    \\ \hline
DSP (SR)              & \ch\checkmark  & \ch\checkmark  & \ch\checkmark     & \ch\checkmark       & \ch\checkmark          & \ch\checkmark          & \ch\checkmark          & \ch\checkmark   & \ch\checkmark     & \ch\checkmark    \\ \hline
DSP (CS+SR)         & \ch\checkmark  & \ch\checkmark  & \ch\checkmark     & \ch\checkmark       & \ch\checkmark          & \ch\checkmark          & \ch\checkmark          & \ch\checkmark   & \ch\checkmark     & \ch\checkmark    \\ \hline
\end{tabular}
\caption{Round-level success rates on 10 existing poker variants for 20 rounds. We use {\ch\checkmark} to indicate the 100\% success rate. CS and SR refer to the core set and segment rephrasing technique.}
\label{tab:in-d}
\end{table*}

\begin{table*}[]
\centering
\small
\setlength\tabcolsep{4.2pt}
\begin{tabular}{|l|c|c|c|c|c|c|c|c|c|}
\hline
                 & \texttt{start} (A) & \texttt{blind} (C) & \texttt{shuf.} (D) & \texttt{deal} (C) & \texttt{flop} (C) & \texttt{switch} (B) & \texttt{bet} (B)  & \texttt{show} (A) & \texttt{prize} (A) \\ \hline
GPT4 (5-shot)    & 88.0           & 84.0           & \ch\checkmark    & 31.3          & 77.6          & 20.6            & 78.7          & 0.0           & 83.0          \\ \hline
CoGem. (5k)      & 94.0           & \ch\checkmark  & \ch\checkmark    & \ch\checkmark & \ch\checkmark & \ch\checkmark   & 93.0          & 88.0          & \ch\checkmark \\ \hline
CoGem. (10k)     & \ch\checkmark  & \ch\checkmark  & \ch\checkmark    & \ch\checkmark & \ch\checkmark & \ch\checkmark   & \ch\checkmark & \ch\checkmark & \ch\checkmark \\ \hline
\end{tabular}
\caption{State-level performances with different values of training data based CodeGemma-7b (DSP+CS+SR). We use {\ch\checkmark} to indicate the 100\% accuracy. We label the difficulty of each type of states from A to D from hard to easy.}
\label{tab:in-d-flow}
\end{table*}

$\bullet$ \textbf{In-domain evaluation}: The model has been exposed to a broad range of variants based on ten existing poker games during training. We sampled some unseen variants of these ten games from the poker simulator for evaluation. Then, we program some random players that randomly select an action as their input to interact with the IDGE. This manner allows for a quick and automatic assessment of the IDGE's basic performance as well as the effectiveness of training methods. Specifically, each type of games is played for 20 rounds. There are totally 200 rounds of games in the in-domain test set.
The state prediction accuracy is determined through two steps. First, we compare the predicted code snippet and the ground truth. If not exactly matched, we execute both snippets on the input state respectively and then compare two outputs.

$\bullet$ \textbf{Out-of-domain evaluation}: The in-domain evaluation is limited to a number of predefined poker games with configurable essential elements. To evaluate our the IDGE's performance in scenarios more closely aligned with the real world, we further recruit 5 proficient poker players as our engine testers. Each of them is asked to create 1$\sim$2 new poker games based on their personal preferences and craft the game script using natural language. They are free to tailor the game scripts, for example, crafting the entirely new elements and strategies not found in existing poker games. Subsequently, we invite them to play 10 rounds of the game with distinct configurations for each new game by themselves and record all player inputs and game states throughout the game-play. This forms our out-of-domain test set that comprises 8 distinct game scripts and 80 rounds of games.

\subsection{In-Domain}
\noindent\textbf{Round-level}\;
Table \ref{tab:in-d} shows the round-level success rates of a number of fine-tuned models. The success rate is counted if the engine correctly handles all states in a round.
The results of CodeGemma from NSP to DSP suggest the advantage of predicting the difference of two states, which results in both accuracy and efficiency boost.
The best results occur when the model undergoes segment rephrasing (SR) and the full curriculum (CS + SR) respectively. The resultant CodeGemma achieves 100\% success rates on all ten poker variants. This suggests the effectiveness of SR to enhance the model's understanding on the game scripts.
In the following, we will show that SR is more important in the face of out-of-domain games.

\noindent\textbf{State-level}\;
We also introduce GPT4 as a strong baseline in our experiment, which is prompted with additionally five in-context samples (5-shot). Surprisingly, in 200 rounds of games, it is unable to successfully complete any single round.
One might question why GPT4 completely fails in this task, significantly behind fine-tuned CodeGemma-7b.
To conduct a more in-depth analysis, we compute the state-level accuracy in Table~\ref{tab:in-d-flow}.
We find that, though GPT4 is strong in programming, it performs badly in managing nuanced poker cards.
For example, it is very likely to mess up the order, hallucinating new cards or missing some of them. This drawback is pronounced in \texttt{deal} and \texttt{show}. In contrast, \texttt{deal} is a much easier task for humans.
We conjecture that current LLMs have not been exposed to highly sophisticated data and tasks as for the IDGEs during their training.
The accumulation of errors in all these aspects eventually leads non-fine-tuned models to zero success rates in round-level evaluation.
It is important to note that an IDGE should be all-round at each aspect; otherwise, the overall performance will degenerate in a way of \textbf{Buckets effect}.

In contrast, for fine-tuned CodeGemma, Table~\ref{tab:in-d-flow} shows that it has performed close to 100\% accuracy in most states with only a half of training samples (5k). Such high accuracy correlates positively to its stable round-level performance in Table \ref{tab:in-d}.
We notice that CS is particularly beneficial for \texttt{show}, where the model is responsible to calculate the hand combinations and compare their strength, the most challenging task in poker games. There are a large number of relevant core functions in this process. Hence, it becomes critical for the model to adapt to core functions in advance.

\subsection{Out-of-Domain}

\begin{table}[]
\small
\centering
\begin{tabular}{lcc}
\toprule
                        & IDGE           & IDGE w. SR    \\
\textsc{Magic Dealer}   & \x\XSolidBrush & \ch\checkmark \\
\textsc{3-card Draw}    & 8/10           & \ch\checkmark \\
\textsc{6-card Draw}    & \x\XSolidBrush & \ch\checkmark \\
\textsc{Dragonie}       & 1/10           & 9/10          \\
\textsc{Three Kingdoms} & \x\XSolidBrush & \ch\checkmark \\
\textsc{Stardust}       & \x\XSolidBrush & 8/10          \\
\textsc{Odd Lover}      & 3/10           & \ch\checkmark \\
\textsc{Joker Hold'em}  & \x\XSolidBrush & \ch\checkmark \\ \bottomrule
\end{tabular}
\caption{Success rates on out-of-domain games.}
\label{tab:oo-d}
\end{table}

Table~\ref{tab:oo-d-script} in Appendix \ref{apx:2} illustrates the eight scripts created by human players. Most of them are creative new games with a large gap from standard poker. For example, in script 6, the creator defines a group of novel combinations ``Stardust X''.

Table \ref{tab:oo-d} reports the round-level success rates of our IDGE, fine-tuned based on CodeGemma-7b with and without SR.
We first find that the model not underwent SR fails to be fully instructable by players.
For example, it cannot understand the tricky dealing process in \textit{Magic Dealer} described in free natural language, though it is a simple variant from standard dealing.
In contrast, the model underwent SR treats this with ease. The rephrased samples encourage the model to learn the alignment between prototype game scripts and diverse natural language, thereby better balancing stability and diversity.
Additionally, the full IDGE demonstrates remarkable generalizability in the face of novel and unseen games.
For example, in \textit{6-card Draw}, the IDGE effectively generalizes from managing 5-card hands to 6-card hands, while in \textit{Dragonie}, which is an upgrade version of \textit{Badugi}, the IDGE learns to pick out cards with distinct suits while determining the consecutiveness of their values.
For more challenging \textit{Stardust}, where the creator introduces a series of entirely new cards and combinations, the IDGE successfully passes eight of the ten rounds of the game.

\section{Conclusion}
This paper introduces the Instruction-Driven Game Engine (IDGE), offering game enthusiasts a brand new game development and game-play experience. The IDGE understands the player-specified game rules and simulates the entire game-play process. We formulate the learning of IDGEs as Next State Prediction and leverage a curriculum learning approach to enhance stability and diversity. Experiments demonstrate our poker IDGE can accurately complete the majority of user-defined games.

\section*{Broader Impact}
This paper presents the initial progress of IDGE in the case of Poker. Such a paradigm theoretically applies to all types of games. However, our progress is constrained by several bottlenecks.

\noindent\textbf{Inference Latency} We have demonstrated that IDGEs go well with turn-based strategy (TBS) games. For real-time strategy (RTS) games, players may make more than one action per second. The inference latency of current LLMs cannot meet the real-time requirements of such games.

\noindent\textbf{Context Window} Generally, as games become more complicated, the length of game states increases, posing a challenge to satisfy our independence assumption. This may significantly challenge both the comprehension ability of LLMs and the cache of KV states.

\noindent\textbf{Accessibility} The kernel data of most commercial games is not publicly available, which is why we developed a poker simulator to generate the training data for this paper.

We are delighted to observe that there have been continuous advancements in inference frameworks such as vLLM \citep{kwon2023efficient}, as well as efficient long-text generation methods like StreamingLLM \citep{DBLP:conf/iclr/XiaoTCHL24} and Temp-LoRA \citep{DBLP:journals/corr/abs-2401-11504}. We believe that the ongoing development of LLM technologies will ultimately address the limitations of latency and the context window. Regarding the issue of accessibility, we look forward to more companies providing open interfaces as SC2LE \citep{DBLP:journals/corr/abs-1708-04782}, HOK Arena \citep{DBLP:conf/nips/0001CJQDL000LHY22} to offer kernel data.

The recent released Delta-Engine \citep{DBLP:journals/corr/abs-2408-05842} is largely inspired from our work. It exclusively focuses on game development. The development process can be ideally eternal, by expanding the engine incrementally. Unlike the IDGE, the delta-engine does not simulate the game-play process. The resultant game-play is rendered by external modules.

% \section*{Acknowledgments}

% This document has been adapted
% by Steven Bethard, Ryan Cotterell and Rui Yan
% from the instructions for earlier ACL and NAACL proceedings, including those for
% ACL 2019 by Douwe Kiela and Ivan Vuli\'{c},
% NAACL 2019 by Stephanie Lukin and Alla Roskovskaya,
% ACL 2018 by Shay Cohen, Kevin Gimpel, and Wei Lu,
% NAACL 2018 by Margaret Mitchell and Stephanie Lukin,
% Bib\TeX{} suggestions for (NA)ACL 2017/2018 from Jason Eisner,
% ACL 2017 by Dan Gildea and Min-Yen Kan,
% NAACL 2017 by Margaret Mitchell,
% ACL 2012 by Maggie Li and Michael White,
% ACL 2010 by Jing-Shin Chang and Philipp Koehn,
% ACL 2008 by Johanna D. Moore, Simone Teufel, James Allan, and Sadaoki Furui,
% ACL 2005 by Hwee Tou Ng and Kemal Oflazer,
% ACL 2002 by Eugene Charniak and Dekang Lin,
% and earlier ACL and EACL formats written by several people, including
% John Chen, Henry S. Thompson and Donald Walker.
% Additional elements were taken from the formatting instructions of the \emph{International Joint Conference on Artificial Intelligence} and the \emph{Conference on Computer Vision and Pattern Recognition}.

\bibliography{custom}

\newpage
\appendix

\section{Related Work}
A game engine is a fundamental software designed for game development.
Famous game engines include Unreal, Unity, CoCos, etc. Pygame is also a simple game engine.
We spotlight two crucial properties of a game engine.
The first is functionality, i.e. providing a wide variety of basic tools to facilitate the development process.
The next is secondary development, i.e. rich and flexible interfaces to allow developers to customize games.
In this work, we introduce a new concept, instruction-driven game engine (IDGE), a neural game engine learned on basis of large language models \citep{DBLP:journals/corr/abs-2303-08774,DBLP:journals/corr/abs-2307-09288,DBLP:journals/corr/abs-2310-06825,DBLP:journals/corr/abs-2309-10305,DBLP:journals/corr/abs-2304-08354}.
As opposed to typical game engines, the IDGE acquires its functionality power by instruction tuning on the core set \citep{DBLP:journals/jmlr/RaffelSRLNMZLL20,DBLP:conf/nips/Ouyang0JAWMZASR22} and allows for low-barrier game development by issuing natural language descriptions.

Some research efforts have explored the AI applications in games \citep{DBLP:journals/corr/abs-2402-18659}, e.g. non-play characters \citep{DBLP:journals/nature/ShanahanMR23,DBLP:journals/air/UludagliO23}, interactive drama \citep{DBLP:conf/acl/WuWJL0024,DBLP:conf/acl/HanCLXY24}, game commentators \citep{DBLP:conf/icids/Eladhari18,DBLP:conf/exag/RanellaE23}.
A great amount of work focuses on AI as players, e.g. for Atari \citep{DBLP:journals/corr/MnihKSGAWR13}, Minecraft \citep{DBLP:conf/nips/FanWJMYZTHZA22,DBLP:journals/corr/abs-2305-16291}, StarCraft \citep{DBLP:journals/nature/VinyalsBCMDCCPE19}, NetHack \citep{DBLP:conf/nips/KuttlerNMRSGR20,DBLP:conf/iclr/Lowe0FKP20}, Werewolf \citep{DBLP:journals/corr/abs-2309-04658}; 
However, our work diverges from all of them in that we treat AI as the playground, attempting to build a game engine that is defined by instructions (game scripts) and game states.
The former focuses on the way AI behaves, while the latter focuses on the way AI would react in the face of any possible behaviors from human beings and agents.
More recent work comes up with learning for a foundation agent, a single agent with generalizable skills to behave in various environments, e.g. SIMA \cite{DBLP:journals/corr/abs-2404-10179}, an instruction-driven agent proficient in multiple simulated environments; CRADLE \cite{DBLP:journals/corr/abs-2403-03186}, a powerful agent capable of playing complex AAA games like Red Dead Redemption 2 by controlling the keyboard and mouse.
However, our work targets the IDGE for a specific group of games, Poker, as an initial step for building a foundation IDGE.
Poker is a widely studied information game of immense popularity \citep{DBLP:journals/cacm/BowlingBJT17,DBLP:journals/corr/MoravcikSBLMBDW17,DBLP:journals/corr/abs-2308-12466,DBLP:journals/corr/abs-2308-07327,DBLP:conf/aaai/ZhaoYLLX22}.

In this paper, the entire training cycle for IDGE is a way of curriculum learning \citep{DBLP:conf/icml/BengioLCW09}. Recent studies show the potential of curriculum learning in empowering the language models to tackle more challenging tasks \citep{DBLP:conf/emnlp/VakilA23,DBLP:conf/emnlp/Wu0ZZ23}.
The proposed segment rephrasing technique is related to perturbation training \citep{DBLP:conf/iclr/ZhuCGSGL20,DBLP:conf/iclr/WuLSZZ23}, which smooths the structured natural language in the semantic space.

\section{Out-of-Domain Game Scripts}
\label{apx:2}

\sethlcolor{lightdelectricblue}
\begin{table*}[]
\centering
\small
\begin{tabular}{@{}l@{}}
\toprule[1.1pt]
\multicolumn{1}{c}{Script 1: \textsc{Magic Dealer}} \\ \midrule[1.1pt]
\begin{tabular}[c]{@{}l@{}}The game proceeds in the following order: start the game, shuffling, set blinds, deal 2 cards, bet, reveal 3 cards (the flop), bet,\\ reveal 1 card (the turn), \hl{deal 1 card (new deal)}, bet, show, and finally the prize is distributed. In each dealing phrase, \hl{deal x+1 cards}\\ \hl{to each player. Then randomly discard 1 card from each player's hand and shuffle it back into the deck.} In each flop, flop x cards\\ from the deck to the community. Except when x$=$1, \hl{if the first flopped card matches the suit of the last flopped card, flop 1 more.} \end{tabular} \\ \midrule[1.1pt]
\multicolumn{1}{c}{Script 2: \textsc{3-card Draw}} \\ \midrule[1.1pt]
\begin{tabular}[c]{@{}l@{}}Introduce a new game, named ``3-card draw''. In this game, there are 3 suits, H, D, C, and each player is dealt with a 3-card hand.\\
There are 6 possible combinations of hand. Pair: Two cards of the same value; Three of a Kind: Three cards of the same value;\\
\hl{Straight: Three consecutive cards of any suit;} \hl{Flush: Three cards of the same suit, not consecutive;} Straight Flush: Three consecutive\\ cards of the same suit; High Card: Not conforming to any of above combinations. \end{tabular} \\ \midrule[1.1pt]
\multicolumn{1}{c}{Script 3: \textsc{6-card Draw}} \\ \midrule[1.1pt]
\begin{tabular}[c]{@{}l@{}}Introduce a new game ``6-card draw''. In this game, there are four suits, Hearts (H), Diamonds (D), Clubs (C), Spades (S).\\
In addition, define \hl{two new combinations with 6 cards in hand.}\\
\hl{Three Pair}: there are three pairs of distinct numbers, e.g. D8, H8, C10, H10, H12, D12.\\
\hl{Big House}: there are two pairs of three of one kind, e.g. H8, C8, S8, C12, H12, D12.\\
All combinations rank as: High Card$<$Pair$<$Three of a Kind$<$Straight$<$Flush$<$\hl{Full House$<$Three Pair$<$Big House}$<$Straight Flush. \end{tabular} \\ \midrule[1.1pt]
\multicolumn{1}{c}{Script 4: \textsc{Dragonie}} \\ \midrule[1.1pt]
\begin{tabular}[c]{@{}l@{}} There are four original suits: Hearts (H), Diamonds (D), Clubs (C), Spades (S). \hl{There is an additional superior suit: Loong (L).}\\
\hl{The suits rank as: L$>$H$=$D$=$C$=$S.} Card values rank as: 1$<$2$<$3$<$4$<$5$<$6$<$7$<$8$<$9$<$10$<$11$<$12$<$13.\\
Introduce a new ranking strategy: ``Dragonie''. For each player with four hole cards, pick out the consecutive cards of distinct suits.\\
\hl{Dragonie refers to the four-card hand where four cards are of consecutive cards as well as of distinct suits.} In this case,\\ the valid cards are four.
In the case that there are three consecutive cards of distinct suits, the valid cards are three.\\
Dragonie$>$three valid cards$>$two valid cards$>$one valid cards. To compare the same number of valid cards, the lowest one is the best. \end{tabular} \\ \midrule[1.1pt]
\multicolumn{1}{c}{Script 5: \textsc{Three Kingdoms}} \\ \midrule[1.1pt]
\begin{tabular}[c]{@{}l@{}} These new poker game is called ``Three Kingdoms''. \hl{There are three distinct suits: Shu Han (S), Cao Wei (W), and Dong Wu (D).} \\
Each player will be dealt with four hole cards. \hl{The biggest hand is the one where at least one of all three kingdoms (suits) is present,} \\
\hl{call it ``Three Kingdoms''.} The second biggest is the one where at least two kingdoms is present, ``Two Kingdoms''. The rest of the \\situations belong to Hard Card. In these game, highest cards are preferred when comparing two hands of the same combination. \end{tabular} \\ \midrule[1.1pt]
\multicolumn{1}{c}{Script 6: \textsc{Stardust}} \\ \midrule[1.1pt]
\begin{tabular}[c]{@{}l@{}}\hl{There are ten special cards: ``Stardust''} in the deck (represented as *). These cards are of none suit and none value.\\
\hl{In hand with one Stardust card, the required number of cards to form a straight or flush will be one less, and is greater than a normal}\\ \hl{straight or flush.} The hand with more than one Stardust, will be reduced to High Card. In detail,\\
\hl{Stardust Straight: Four consecutive cards of any suit, plus a Stardust (*);}\\
\hl{Stardust Flush: Four cards of the same suit, not consecutive, plus a Stardust (*);}\\
\hl{Stardust Straight Flush: Four consecutive cards of the same suit, plus a Stardust (*).}\\
High Card$<$Pair$<$Three of a Kind$<$Straight$<$Stardust Straight$<$Flush$<$Stardust Flush$<$Straight Flush$<$Stardust Straight Flush. \end{tabular} \\ \midrule[1.1pt]
\multicolumn{1}{c}{Script 7: \textsc{Odd Lover}} \\ \midrule[1.1pt]
\begin{tabular}[c]{@{}l@{}} In this game, odd values (1, 3, 5, 7, 9) are greater than even values (2, 4, 6, 8, 10). They rank as: \hl{2$<$4$<$6$<$8$<$10$<$1$<$3$<$5$<$7$<$9.} \\
Card combinations rank as: High Card$<$Odd Straight$<$Odd Flush$<$Odd Straight Flush. \\
\hl{Odd Straight}: Five consecutive odd values of any suit, e.g. 1, 3, 5, 7, 9; \hl{Odd Flush}: Five odd values of the same suit, not consecutive; \\
Odd Straight Flush: Five consecutive odd values of the same suit; High Card: Not conforming to any of above combinations. \end{tabular} \\ \midrule[1.1pt]
\multicolumn{1}{c}{Script 8: \textsc{Joker Hold'em}} \\ \midrule[1.1pt]
\begin{tabular}[c]{@{}l@{}}There are four suits: Hearts (H), Diamonds (D), Clubs (C), Spades (S). Card values rank as: 2$<$3$<$4$<$5$<$6$<$7$<$8$<$9$<$10$<$J$<$Q$<$K$<$1.\\
\hl{In addition, there are two special Joker cards represented as J1 and J2, which can be treated as any suit and value.} \\
Three of a Kind: Three cards of the same value. Straight: Five consecutive cards of any suit. Flush: Five cards of the same suit, not \\ consecutive.
Full House: Three cards of the same value plus another two cards of another value. Four of a Kind: Four cards of the \\ same value.
\hl{Five of a Kind: Five cards of the same value (possibly with Joker).} Straight Flush: Five consecutive cards of the same suit. \\ High Card: Not conforming to any of above combinations. \end{tabular} \\
\bottomrule
\end{tabular}
\caption{Out-of-domain game scripts written by human players. We skip some basic settings in the script for brevity, e.g. the number of players, bet limits.}
\label{tab:oo-d-script}
\end{table*}

\section{System Demonstration}
\label{apx:3}

\begin{figure}[b!]
\centering
\includegraphics[width=0.43\textwidth]{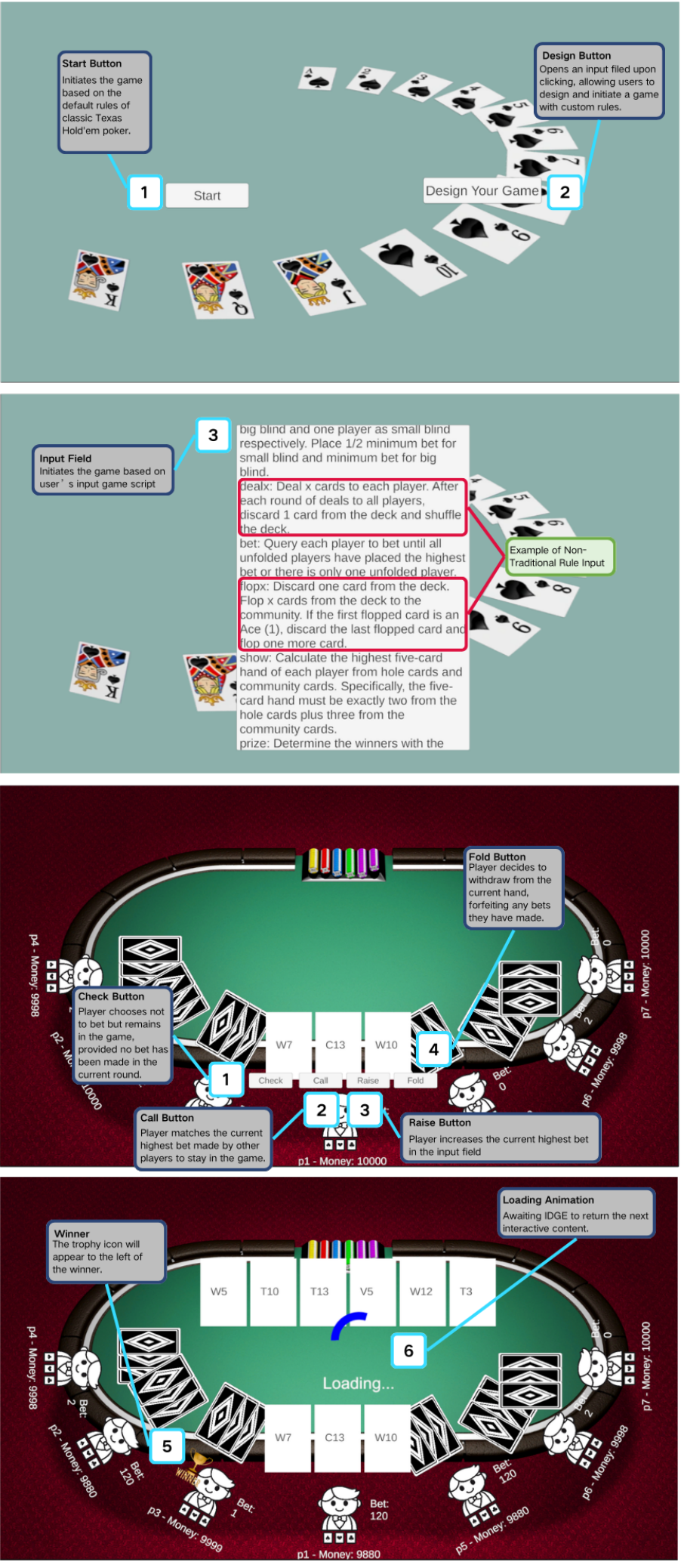}
\caption{System demonstration of our poker IDGE, developed based on Unity.}
\label{fig:unity}
\end{figure}

\end{document}